# A Temporally and Spatially Local Spike-based Backpropagation Algorithm to Enable Training in Hardware

Anmol Biswas[1], Vivek Saraswat, Udayan Ganguly

*Electrical Engineering Dept., Indian Institute of Technology, Bombay*

**Abstract**

Spiking Neural Networks (SNNs) have emerged as a hardware efficient architecture for classification tasks. The challenge of spike-based encoding has been the lack of a universal training mechanism performed entirely using spikes. There have been several attempts to adopt the powerful backpropagation (BP) technique used in non-spiking artificial neural networks (ANN): (1) SNNs can be trained by externally computed numerical gradients. (2) A major advancement towards native spike-based learning has been the use of approximate Backpropagation using spike-time dependent plasticity (STDP) with phased forward/backward passes. However, the transfer of information between such phases for gradient and weight update calculation necessitates external memory and computational access. This is a challenge for standard neuromorphic hardware implementations. In this paper, we propose a stochastic SNN based Back-Prop (SSNN-BP) algorithm that utilizes a composite neuron to *simultaneously* compute the forward pass activations and backward pass gradients explicitly with spikes. Although signed gradient values are a challenge for spike-based representation, we tackle this by splitting the gradient signal into positive and negative streams. The composite neuron encodes information in the form of stochastic spike-trains and converts Backpropagation weight updates

*Email addresses:* `194076019@iitb.ac.in` (Anmol Biswas), `svivek@iitb.ac.in` (Vivek Saraswat), `udayan@ee.iitb.ac.in` (Udayan Ganguly)

[1]Corresponding Author

into temporally and spatially local spike coincidence updates compatible with hardware-friendly Resistive Processing Units (RPUs). Furthermore, we characterize the quantization effect of discrete spike-based weight update to show that our method approaches BP ANN baseline with sufficiently long spike-trains. Finally, we show that the well-performing softmax cross-entropy loss function can be implemented through inhibitory lateral connections enforcing a Winner Take All (WTA) rule. Our SNN with a 2-layer network shows excellent generalization through comparable performance to ANNs with equivalent architecture and regularization parameters on static image datasets like MNIST, Fashion-MNIST, Extended MNIST, and temporally encoded image datasets like Neuromorphic MNIST datasets. Thus, SSNN-BP enables BP compatible with purely spike-based neuromorphic hardware.

## 1. Introduction

Spiking Neural Networks have received long-standing interest due to their biological plausibility, efficiency as a result of the discrete nature of information propagation in such networks, and theoretical computation power[1]. Additionally, recent developments of the event-driven neuromorphic hardware [2],[3] have led to a rise in interest in realizing large-scale SNNs. SNNs implemented on such hardware are highly power-efficient compared to current neural networks which require GPUs to perform the matrix operations during learning and inference. Gradient backpropagation [4] is the basic algorithm that reliably tunes the weights and biases of non-spiking artificial neural networks and is responsible for neural networks becoming the foremost approach to solving pattern recognition and classification tasks. However, it is not straightforward to implement Backpropagation in Spiking Networks because Spiking Networks perform computations through discrete spike events that are essentially non-differentiable. Several approaches have been proposed to address this fundamental problem.

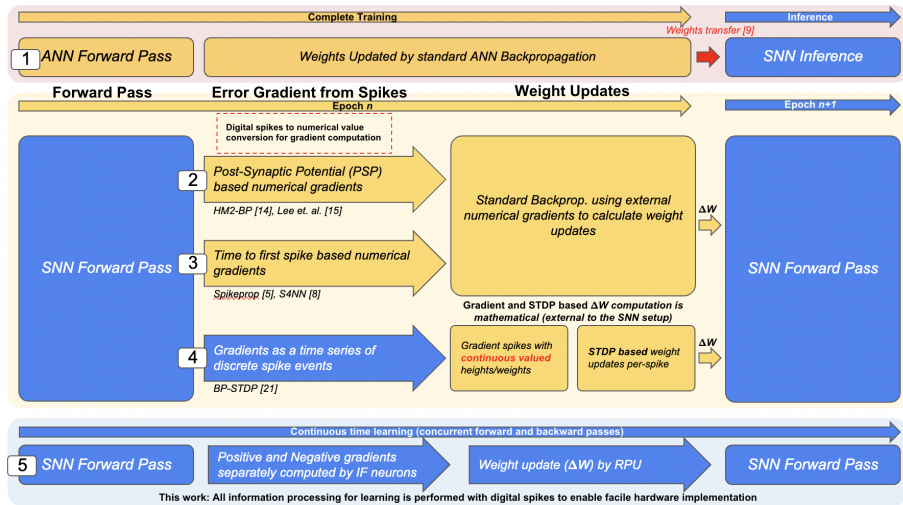

Figure 1: A comparison of the proposed algorithm with existing approaches towards learning in spiking networks. *Blue* represents purely spike-based processing while *Yellow/Orange* represents numerical processing external to the spike-based architecture. Previous work relies on generating numerical values for activations, gradients, and updates in a non-spiking manner using external computation. We propose a continuous time learning algorithm where all operations in backpropagation are performed with spiking neurons.

One of the earliest approaches to attempt learning in Spiking Networks was SpikeProp [5], which minimized the distance between a single target and output spike by formulating the time-to-first-spike as a function of the input to the neuron. This approach was later extended to multiple spike outputs in [6] and [7]. However, time-to-first-spike approaches have fallen out of favor as they have been unable to compete with the performance of more recent approaches due to the weak dependence of the *time to first spike* on the input spike-train compared to the dependence of the *rate of spiking* on the same input. This holds even for the latest in time to first spike algorithms, such as S4NN[8].

An alternative and highly successful approach involves adapting trained Artificial Neural Networks (ANNs) to Spiking Networks [9],[10] and [11] where the ANN is trained separately and then the network weights are transferred to the SNN. The obvious drawback of such an approach however is that it rules out the possibility of online (or on-device) learning entirely and the training task does not benefit from the power-efficient encoding that SNNs have to offer.

Finally, the third broad approach is to adapt the Backpropagation algorithm itself to rate-coded Spiking Networks. This can be done in a number of ways at various levels of complexity. In GD4SNN[12] and SLAYER[13], the instant of spiking is approximated as a continuous-time event whereas HM2-BP[14], [15] and [16] utilize exact Post-Synaptic Potentials (PSPs), to externally compute the Backpropagation gradients with high accuracy. LSNN [17], ST-RSBP [18], e-prop[19] and OSTL[20] utilizes Backpropagation Through Time (BPTT) with PSPs or pseudo-derivatives to train Recurrent Spiking Neural Networks. These methods show state-of-the-art performance in classification problems using SNNs. *However, one aspect common to all of these approaches is that the gradient signals need to be numerically calculated, as shown in Fig. 1, externally to the spiking network.* This limits the plausible implementation of these learning algorithms in event-driven hardware because the computations involved cannot be simply translated to counting and/or timing spiking events.

In contrast, BP-STDP[21] and EMSTDP[22] enables spike-time dependent weight update. However, they are phased, serial algorithms where the backward pass is performed after the forward pass is completed and the forward pass spike trains have to be stored externally. For BP-STDP, during the backward pass, the forward pass spike train is used to implicitly compute the loss gradient at each spike as a signed analog value externally as opposed to within the network. The weight updates occur for each pre-neuron, and post-neuron spike through "approximate" STDP/Anti-STDP rules where the instantaneous gradient is used. Similarly, EMSTDP uses specialised units called trace counters to enable weight updates and several pre-trained convolutional layers. Thus in these approaches, separate forward and backward passes are used where information must be stored and transferred between the forward and backward pass phases externally. This process makes learning temporally non-local and requires significant external memory and computation.

Thus, it is an unsolved challenge to enable temporally local updates for a robust supervised algorithm using only conventional SNN elements without requiring external memory or numerical computation. This requires *all* the computation to be done using *only* neurons and synapses in *concurrent* forward & backward passes.

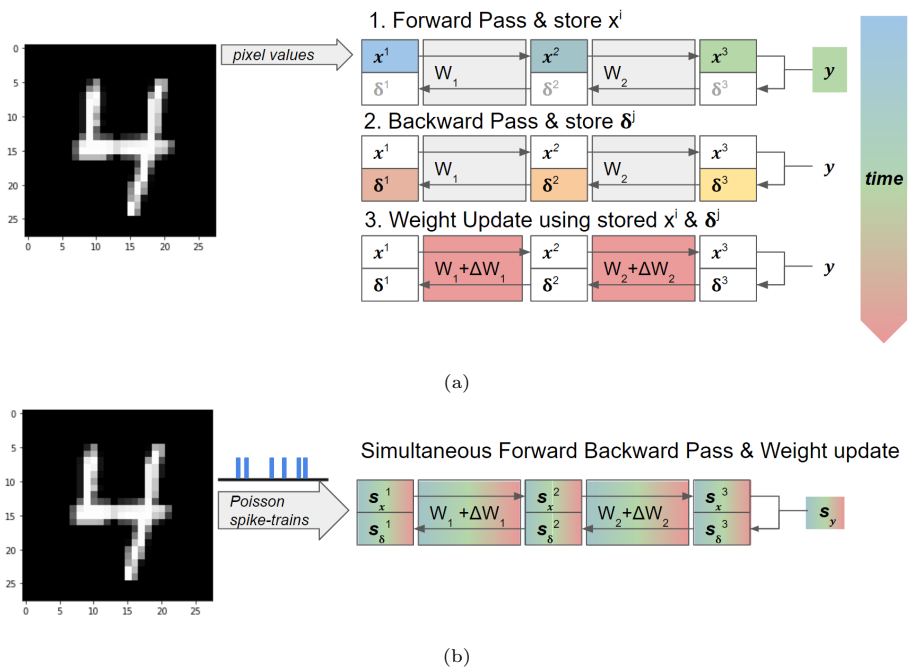

(a)

(b)

Figure 2: Network Architecture and pipeline **(a)** A typical serial pipeline for learning in ANNs and SNNs **(b)** Our learning algorithm with concurrent rate-coded computation of forward and backward passes which enables spatially and temporally local weight updates, making external storage and computations unnecessary

In this paper, we propose a fully spike-based algorithm to compute all Backpropagation gradient spiking signals for training SNNs and a temporally local weight updates scheme for in-memory on-chip learning inspired by Resistive Processing Units (RPUs)[23]. First, our algorithm computes both forward-pass spikes and backpropagation gradient spikes concurrently with the incremental updates which eliminates additional memory requirements. We maintain mathematical equivalence of the spike-domain learning algorithm with training in conventional non-spiking ANNs by noting that the spiking frequency of an integrate and fire (IF) neuron resembles ReLU like [24] characteristics (linear after a threshold). Second, although compartmentalized neurons for forward and backward passes have been reported earlier (SpikeGrad) [25], we enable positive and negative weight updates with unsigned spikes by splitting the gradient compartment into separate positive and negative spiking streams. Third, we propose lateral inhibitory connections in the output layer to implement a winner-take-all rule[26] in order to approximate the softmax cross-entropy loss function[27]. Fourth, we adapt our algorithm to use the resistive processing unit (RPU) concept to compute weight update in a temporally and spatially local manner which eliminates the need for external computation of weight updates.

Fig. 1 illustrates the difference between existing approaches for spike-based training and the proposed algorithm and Fig. 2 visualizes the simultaneous computation. We call our method *Stochastic SNN-BackProp* (SSNN-BP) as it requires stochastic encoding of the input and target spikes for the proposed update algorithm (RPU-inspired temporally local updates) to work. Finally, we show the equivalence of RPU-inspired temporally local updates to a quantized rate-based updates approach. The latter is faster to implement in software and helps us quickly predict the performance of the proposed RPU updates algorithm for different data sets i.e. MNIST, Extended MNIST, and Fashion

MNIST to demonstrate generalization. The rate-based updates approach also helps us estimate the number of simulation steps ($T_s$) required to achieve a target level of performance for the proposed temporally local RPU-inspired updates based on the precision (quantization levels) of the weight updates.

## 2. Proposed Method - SSNN-BP

*2.1. Composite Spiking Neurons*

We propose a composite spiking neuron that can simultaneously compute both the feedforward spikes and the backpropagated gradient spikes with positive and negative streams as the fundamental unit of our network. Further, we introduce an approximation of the softmax cross-entropy loss implemented using inhibitory lateral connections to enforce winner-take-all (WTA) in the output layer. For stable learning and regularization in our experiments, we also adopt the He weight initialization scheme[28] and dropout[29] respectively.

We use a **Spike Response kernel** method[14] to perform efficient GPU-compatible simulations of our spiking network. In this method, the incoming spike trains at any neuron are convolved with a spike response kernel, which converts weighted discrete delta-function spikes into their corresponding post-synaptic potentials and accumulates their values over time. This convolution operation, then, computes the membrane potential of the neuron across the simulation time in one shot and is followed by a thresholding operation to generate the spike train from the computed membrane potential. Due to the one-shot computation of neuron membrane potential in the kernel method, the thresholding operation is slightly different from the standard *issue spike-and-reset* method of simulations that compute the membrane potentials at every time step in sequence. Since the membrane potentials are pre-computed without resetting, we have to instead, increment the threshold after every spike is issued. In doing this, we are making use of the fact that for a spiking threshold of $\theta$, spikes should be issued at the time instances when the pre-computed membrane potential (computed without any spike-issue resetting) crosses $\theta$, $2\theta$, $3\theta$ and so on.

Additionally, since the membrane potential of any neuron is computed in one shot from known input spikes to that neuron and the thresholding and spike-generation operation occur separately, this method cannot directly implement lateral connections, which are required in our final layer. To solve this problem, we have used an iterative approach where at any given iteration, we use the available spikes of the same layer from the previous iteration to compute new membrane potentials, and therefore new spike trains for the current iteration. Eventually, this method generates stable spike patterns after a few iterations

The composite neuron comprises of three compartments: feedforward spikes $s_x^l(t)$, and backward gradient spikes $s_{\delta+}^l(t)$, and $s_{\delta-}^l(t)$. The subscript $x$ refers to forward pass quantities, while the subscript $\delta$ refers to backward pass quantities. We use Integrate and Fire (IF) neurons similar to the ones described in BP-STDP[21] for our network. The forward pass of our network for all layers except the last layer ($N$) is described by the following equations:

$$s_x^{l+1}(t) = f_\theta(W^l \times (\epsilon_x * s_x^l)(t)) \tag{1}$$

$$\epsilon_x(t) = H(t)(1 - e^{-t/\tau_x}) \tag{2}$$

In Eq. 1 and 2, $s_x^l(t)$ represents the spike-train i.e. time-series of the binary spikes in the forward pass direction of the network for the layer $l$, $W^l$ is the weight matrix connecting layer $l$ to layer $l+1$, $\epsilon_x(t)$ is the **Spike Response kernel**, $H()$ represents the standard *Heaviside* function, $*$ represents convolution in time and $f_\theta$ is the threshold function. While generating the spike trains through the $f_\theta$ operation, the starting threshold for firing is set to $V_{th} = \theta$, and an output spike is issued when the membrane potential $(W_l \times (\epsilon * s_x^l)(t))$ exceeds $V_{th}$. After every spike, the threshold is incremented by $\theta$, i.e., $V_{th} \to V_{th} + \theta$.

The output layer (layer $N$) includes lateral inhibition to implement a winner-take-all scheme in order to approximate the softmax activation function. For this, we adopt an iterative approach (since lateral connections cannot be exactly modeled in a spike response kernel implementation):

$$s_x^N(t)_0 = f_\theta(W^{N-1} \times (\epsilon_x * s_x^{N-1})(t)) \tag{3}$$

$$s_x^N(t)_{k+1} = f_\theta(W^{N-1} \times (\epsilon_x * s_x^{N-1})(t) - W_{inh} \times (\epsilon_x * s_{x\ k}^N)(t)) \tag{4}$$

We iterate over $k$ to converge to a $s_x^N(t)$. All diagonal elements in inhibitory weights $W_{inh}$ are set to 0 to prevent self-inhibition. $W_{inh}$ are fixed and not learnt.

For inference, we identify the neuron in the output layer with the maximum spiking activity, i.e., $argmax(\sum_t (s_x^N(t)))$ which means the classification output is assigned to the output layer neuron that issues the most spikes. This completes the description of the neurons for computing the feedforward spikes.

We have a set of parallel networks that compute the backpropagation gradient spikes. Two such backpropagation neuron compartments ($s_{\delta+}^l(t)$, and $s_{\delta-}^l(t)$) are required for each feedforward neuron compartment ($s_x^l(t)$) because typically the backpropagated gradient can be either positive or negative while having only positive values is sufficient for feedforward neurons as shown by the success of Rectified Linear Units (ReLUs) [24], [30]. We intend to represent all signals using unsigned unit-valued spikes only. The backpropagation step for any hidden layer can be described by the following set of equations:

$$s_{\delta+}^l(t) = (\sum_t s_x^l(t) > 0) \otimes f_\theta(W^{l^T} \times (\epsilon_\delta * s_{\delta+}^{l+1})(t) - W^{l^T} \times (\epsilon_\delta * s_{\delta-}^{l+1})(t)) \tag{5}$$

$$s_{\delta-}^l(t) = (\sum_t s_x^l(t) > 0) \otimes f_\theta(W^{l^T} \times (\epsilon_\delta * s_{\delta-}^{l+1})(t) - W^{l^T} \times (\epsilon_\delta * s_{\delta+}^{l+1})(t)) \tag{6}$$

These backpropagation equations are derived by exploiting the similarity between rate-coded IF spiking neurons and ReLUs [24]. In Eq. 5 and Eq. 6, $\otimes$ represents element-wise multiplication over the time steps of the spike train and $\epsilon_\delta(t) = H(t)(1 - e^{-t/\tau_\delta})$ is the **Spike Response kernel** for the gradient neuron compartments. The term $\sum_t s_x^l(t) > 0$ estimates the gradient of the forward spikes inspired by the step function gradient of the ReLU activation function. The comparison $\sum_t s_x^l(t) > 0$ flips to 1 for the corresponding neuron in layer $l$ as soon as the first spike is issued by that neuron and maintains that value for the rest of the simulation for a given input sample. This arrangement of one feedforward and two backpropagation compartments is given the name *Composite Spiking Neuron* because it is a complete unit that computes all the signals required for learning. The operation of the composite neuron unit is visualized in Fig 4a. The gradient signals at the output layer $N$ are computed as:

$$s_{\delta+}^N(t) = f_\theta((\epsilon_\delta * s_x^{label})(t) - (\epsilon_\delta * s_x^N)(t)) \qquad (7)$$

$$s_{\delta-}^N(t) = f_\theta((\epsilon_\delta * s_x^N)(t) - (\epsilon_\delta * s_x^{label})(t)) \qquad (8)$$

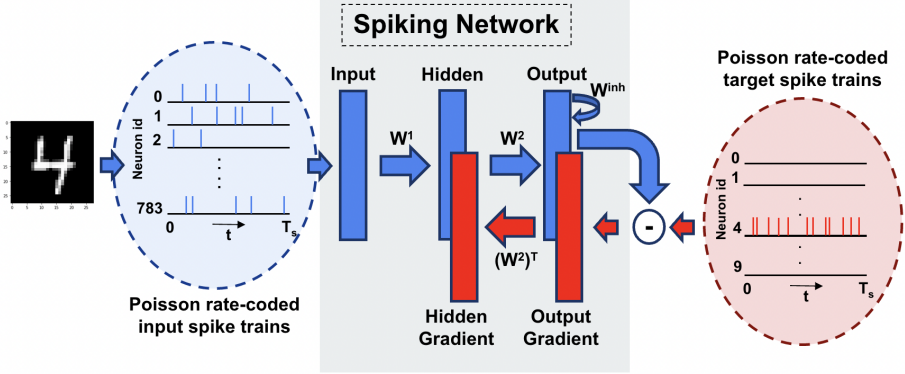

Figure 3: Network Architecture and training. Forward pass compartments are shown in blue and Gradient compartments are shown in red. Inputs and labels are converted to Poisson random spike trains and fed to the spiking network from opposite ends. The forward pass activations and the backpropagation gradients are computed by the respective compartments of the spiking network

Equations Eq. 7 and Eq. 8 are inspired from the common form of the gradient when loss is computed either as mean-squared error or as softmax cross-entropy between $s_x^N(t)$ and $s_x^{label}(t)$, where $s_x^{label}(t)$ is the target output spike train.

The input layer spikes $s_x^1(t)$ and the target spikes $s_x^{label}(t)$ are encoded as stochastic Poisson spike trains of length $T_s$ with average rates derived from input and label values of the dataset (say pixel intensity value for images) as shown in Fig. 3.

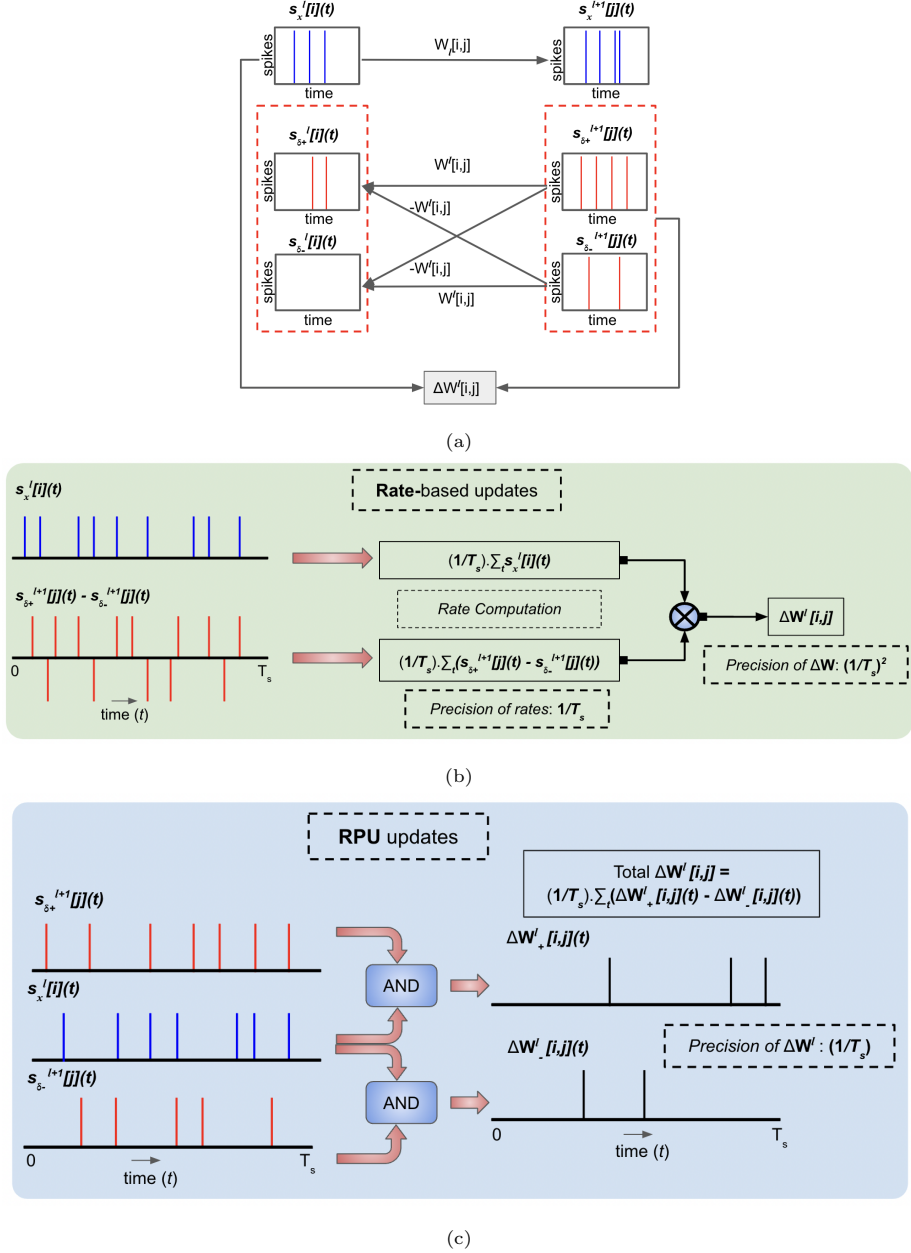

Figure 4: **(a)** Spike-based forward and backward passes and weight updates. The net gradient $s_\delta(t)$ is computed as $(s_{\delta+}(t) - s_{\delta-}(t))$ **(b)** The Rate-based update rule, where spike rates of $s_x^l[i](t)$ and $s_\delta^{l+1}[j](t)$ are computed first before multiplication **(c)** Temporally local weight update rule where the products of the spike rates of $s_x^l[i](t)$ with $s_{\delta+}^{l+1}[j](t)$ and $s_{\delta-}^{l+1}[j](t)$ are computed by counting coinciding spikes (AND function)

### 2.2. RPU-inspired temporally local weight updates

We propose a spatially and temporally local weight update rule inspired by RPU which uses spike coincidence of independent spiking signals for incremental updates. For the method to work, we also need the initial spike trains on both the forward pass and backpropagation directions to be stochastic and independent of each other as shown in Fig. 3

In order to calculate the weight updates for $W^l$, we need the outer product of the activity represented by the signals $s_x^l(t)$ and $s_{\delta+}^{l+1}(t) - s_{\delta-}^{l+1}(t)$. Since we are using independent and stochastic inputs and target spikes for the network, it is possible to implement weight updates as temporally local incremental values. For stochastic and independent spike trains, the coincidence (or "AND") of spikes effectively generates a signal proportional to the product [31]. This is a core feature of RPU-based updates. Further, for binary signals,"AND" is equivalent to a multiplication operation, hence the incremental updates for $ij^{th}$ weight can be written as:

$$\Delta W_{ij,+}^l(t) = \eta/T_s \times (s_{\delta+,j}^{l+1}(t) \times s_{x,i}^l(t)) \tag{9}$$

$$\Delta W_{ij,-}^l(t) = \eta/T_s \times (s_{\delta-,j}^{l+1}(t) \times s_{x,i}^l(t))) \tag{10}$$

Where $\eta$ is the learning rate and $T_s$ is the number of simulation steps per sample. The implementation of incremental updates based on coincidence is another core feature of RPU-based updates. The weights are implemented as an analog accumulative memory with non-linear programming threshold that undergoes an incremental Program/Erase only for overlapping spikes (coincidence) and ignores non-overlapping spikes [23]. Further, the updates can happen concurrently with the forward and backward spikes computations. The overall weight update applied after $T_s$ time steps can be written as:

$$\Delta W_{ij}^l = \eta/T_s \times \sum_t ((s_{\delta+,j}^{l+1}(t) - s_{\delta-,j}^{l+1}(t)) \times s_{x,i}^l(t))) \tag{11}$$

As shown in Fig. 4c, the weight updates computed by this temporally local method have a precision of $1/T_s$, as a result, the updates, and consequently the performance suffers from a quantization issue. As $T_s$ increases, the accuracy is expected to increase. However, it is important to note that the proposed RPU-inspired temporally local update method is aimed at hardware implementations. It is highly inefficient to simulate this stochastic spike coincidence-based update behavior in software especially for high-performing large values of $T_s$. Hence, we propose a strategy to predict the performance of RPU-inspired updates using a quantized rate-based update method as explained next.

2.2.1. *Quantized rate-based weight updates*

The updates to the weight matrix can also be computed as the outer product of $\sum_t s_x^l(t)/T_s$ and $\sum_t (s_{\delta+}^{l+1}(t) - s_{\delta-}^{l+1}(t))/T_s$ in line with backpropagation [4]:

$$\Delta W^l = \eta \times (\sum_t (s_{\delta+}^{l+1}(t) - s_{\delta-}^{l+1}(t))/T_s) \times (\sum_t s_x^l(t)/T_s)^T \qquad (12)$$

Eq. 12 computes the weight update as a product of the rates of two spiking neurons as visualized in Fig. 4b. As shown in Fig. 4b, the weight updates computed by this method with $T_s$ simulation steps have a precision of $1/T_s^2$. Clearly, this update method is not temporally local or purely spike-based (need to calculate average rates), neither is it concurrent with the passes (need to wait for passes to finish to calculate rates and updates). We are introducing this rate-based update method for its capability to predict the performance of the temporally local and concurrent RPU-inspired updates while retaining an efficient software implementation.

Rate-based updates allow us to have a high level of precision in the weight updates with a relatively small value of $T_s$ (taken to be 50 for our baseline simulations, precision becomes 1/2500). In order to estimate the performance for the RPU-inspired method using $T_s$ steps (say 100), we requantize the baseline rate-based updates from high-precision (1/2500) to a low-precision (1/100). In *Section* 3.2, we show that simply quantizing the rate-based weight updates gives us a good way to predict the performance of the RPU-inspired temporally local updates. This enables us to find the number of steps needed by the RPU-inspired updates to achieve a given target accuracy.

Table 1: Simulation Parameters

| Simulation Type | $\tau_x$ | $\tau_\delta$ | $\theta$ | $\eta$ |
|---|---|---|---|---|
| Rate-based, $T_s = 50$ | 5 | 0.5 | 5 | 0.06 |
| RPU-inspired temporally local, $T_s = 100$ | 10 | 1 | 5 | 0.04 |
| RPU-inspired temporally local, $T_s = 200$ | 20 | 2 | 5 | 0.01 |
| RPU-inspired temporally local, $T_s = 300$ | 30 | 3 | 5 | 0.005 |

*2.3. Parameters*

The parameters used for the simulations are provided in table 1. The choice of $\tau$ (neuron time constant) in our algorithm is dictated by the form of encoding that we use. Since we are working with rate-based encoding, we require the neurons in the network to respond to incoming spike rates, while also spiking at a reasonable rate. To achieve this balance, a moderate time constant is used for the forward pass, as it allows the neuron to average incoming spike rates for a reasonable number of timesteps while also not being too slow to raise its membrane potential in response to incoming spikes. A much smaller value is used for the backward pass to increase the number of gradient spikes as very few label spikes are provided as input for the backward pass. The choice of $\tau$ for an arbitrary input spike-train will depend on the spiking statistics of the data and needs to be empirically determined considering the trade-offs explained above. Finally, we would like to clarify that $\tau$ has not been optimized in this work. We have simply used what we believe to be reasonable values to represent spike rate information in the time frames considered for simulation.

The values of $\tau_x$ and $\tau_\delta$ are set to $T_s/10$ and $T_s/100$ respectively, the value of $\theta$ is maintained across simulations and the learning rate $\eta$ corresponding to best learning performance is used. It is found that progressively lower values of $\eta$ are required for *RPU-inspired temporally local* simulations with higher number of simulation steps ($T_s$). We use batch size of 50 for all of our training simulations

## 3. Results

We test our network and supervised learning algorithm on the classification benchmark dataset - MNIST[32]. The equivalence of WTA due to lateral inhibi-tion in the output layer with soft-max cross-entropy loss is demonstrated. The results are compared against the state of the art in the different approaches to learning in SNNs and ANN baseline performance. We characterize the performance of the temporally local RPU-inspired weight updates for different $T_s$, concentrating on the MNIST dataset. We demonstrate the equivalence of the rate-quantized updates to RPU-inspired updates and find the optimal number of time steps needed for target accuracy. Finally, we extend our comparison to more complex benchmark datasets - Fashion-MNIST[33] and EMNIST[34] and show generalizing capability comparable to the state of the art in SNN learning and the ANN baseline as well as the improvements from implementing dropout[29] and approximate softmax cross-entropy[27] loss function in our spike-based learning algorithm. Code: `https://github.com/SNNalgo/SSNN-BP`

### 3.1. Softmax Cross-Entropy Loss Approximation

The softmax cross-entropy loss is an important contributing factor to improving the performance of neural networks in solving difficult classification tasks [27]. Owing to the properties of the Softmax Cross-Entropy Loss function, simply replicating a softmax-like function under a spiking paradigm is sufficient to implement learning with Softmax Cross-Entropy as the loss. However, softmax is a dynamic function that is dependent, not only on the input activation in question but also on all the other activations in the same layer. The most visible effect of the softmax function is the exaggeration of small differences between activations, where the largest activation value is enhanced and everything else is strongly suppressed by comparison. This behavior is qualitatively replicated by the Winner Take All (WTA) networks with inhibitory lateral connections in the output layer[26]. Fig 5 compares the aggregate distribution of our WTA outputs with the corresponding softmax outputs on the same raw spiking output (without inhibitory lateral connections) over the MNIST[32] test set. After normalizing and sorting all outputs by value, the aggregate effect that the softmax function has on the distribution of output layer spiking activity is found to be very similar to the effect of WTA implemented by lateral inhibitory connections.

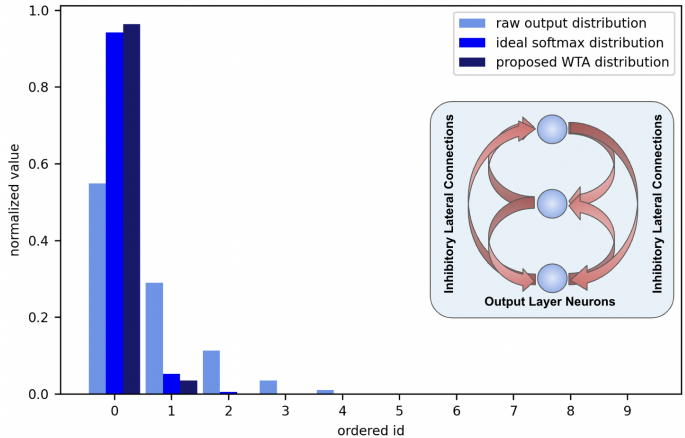

Figure 5: Comparison of the effects of softmax and WTA on the distribution of raw spiking output. Output neuron ids are sorted in descending order of response. While the raw spiking output is not so selective, WTA amplifies the nonlinearity ensuring that the primary neuron spikes uniquely while other neurons are muted. This behavior approximates the softmax function. **Inset**: Inhibitory lateral connections implementing WTA in output layer

*3.2. Predicting performance of temporally local updates with quantization*

As shown in Figs. 4b and 4c, there is a big discrepancy in the precision of the weight updates computed by the rate-based method and the temporally local method. In view of this knowledge, we train an SNN with temporally local weight updates across a range of simulation steps, ($T_s \in \{100, 200, 300\}$) **and** with appropriately quantized baseline rate-based updates ($T_s = 50$) using 30,000 training samples of the MNIST dataset for 50 epochs. We find in our experiments, as shown in Figs. 6 and 7, that simply quantizing the rate-based weight updates with the appropriate number of levels can provide a reliable prediction of the learning performance of the proposed RPU-inspired temporally local weight updates for varying simulation steps ($T_s$). Furthermore, we find that the performances of the quantized rate-based weight updates and temporally local weight updates converge with increasing $T_s$ and increasing number of training epochs. In Fig. 7, we extrapolate to find that $T_s = 1000$ would be sufficient for the temporally local weight updates to match the performance of the baseline rate-based weight updates. In Fig. 8, we extend the training of the SNN with quantized rate-based weight update simulations, both in terms of training data (50,000 training samples) and epochs (200 epochs) and find that the performance difference seen in the shorter and smaller training (50 epochs) for low number of simulation steps, i.e. $T_s = 300$ (Figs. 6, 7) could disappear (become same as $T_s = 1000$) in the long run (200 epochs). After establishing this equivalence, we can now continue to use rate-based update simulations to efficiently predict the performance of the RPU-inspired updates for a large variety of datasets.

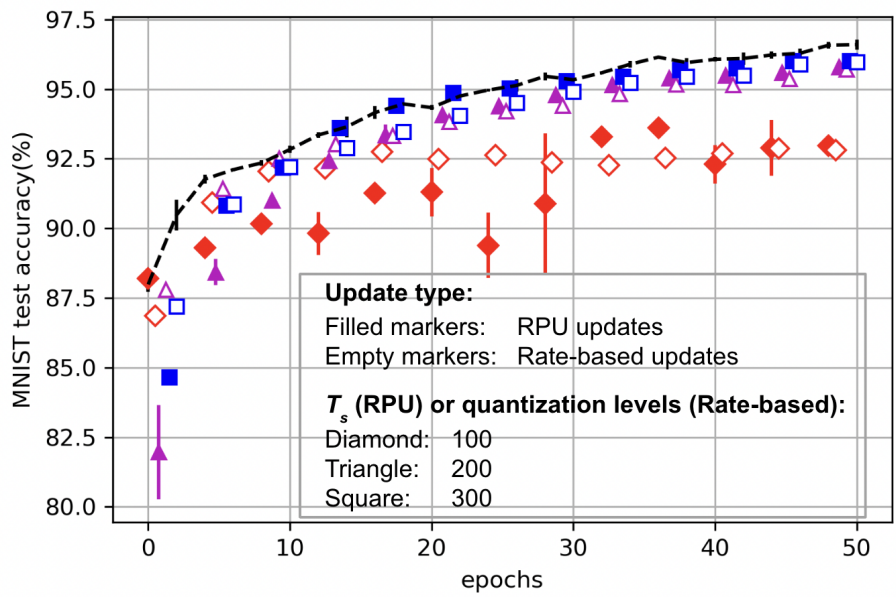

Figure 6: Comparison of learning performance with RPU-inspired temporally local weight updates and quantized baseline Rate-based weight updates. Black dashed line is the full-precision Rate-based baseline

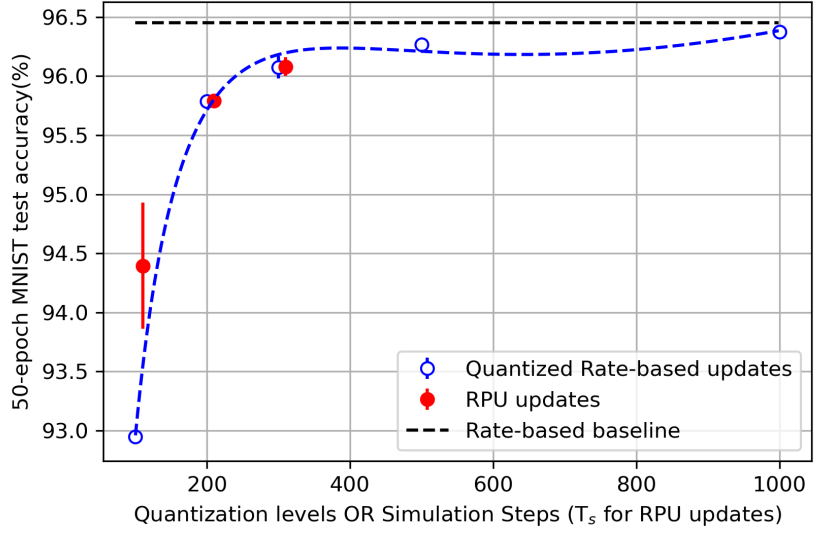

Figure 7: Trend of learning performance of quantized baseline Rate-based weight updates predicting the learning performance of RPU-inspired temporally local weight updates

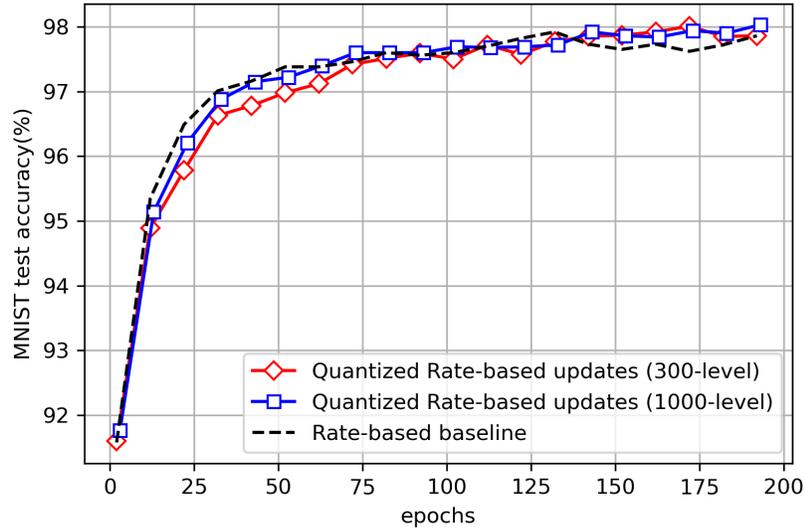

Figure 8: Larger and longer simulations of training with quantized baseline Rate-based weight updates

### 3.3. Performance on different datasets

#### 3.3.1. MNIST

We test the performance of the SNN and the training algorithm on the MNIST[32] handwritten numeric digit classification dataset. The dataset consists of 60,000 training samples and 10,000 test samples. Each sample is a 28x28 grayscale image. The input layer receives randomly generated Poisson spike trains with spike-rate set to the pixel value normalized to $[0, 1]$ and the output layer receives target Poisson spike trains. The neuron corresponding to the correct class receives a target spike train with a spike-rate of 0.5 while the other output layer neurons don't receive any spikes. Training the network with moderate levels of dropout (0.2 in the first layer and 0.3 in the second layer) gives test classification performance of up to 98.65%. The performance of our baseline simulations with Rate-based weight updates is comparable to the state-of-the-art in SNNs [14],[9] as well as the ANN baseline with an equivalent architecture. Table 2 contains detailed comparisons.

#### 3.3.2. Generalization to FMNIST, EMNIST

To showcase the generalizing property of our method, we test it on two additional datasets, *Fashion MNIST* and *Extended MNIST* which are harder classification problems than MNIST despite having identical input dimensions.

Table 2: Comparison of the features and performance of learning algorithms on MNIST dataset

| Model | Network | Loss and Regularizer | Gradient Computation | STDP-like weight updates | Fully SNN Compatible Learning | MNIST test-set accuracy (%) |
|---|---|---|---|---|---|---|
| **BP-STDP** [21] | 3 layer H1=500, H2=150 | MSE | Discrete spikes are used to perform gradient computation and weight update for each spike but using external (non-SNN) mathematical processing | Yes | No | 97.2 |
| Diehl et al. [9] | 3 layer H1=1200, H2=1200 | MSE | Weights are transferred from trained ANN | No | No | 98.64 |
| **HM2-BP** [14] | 2 layer H1=800 | MSE | Numerical gradient computation performed on continuous valued Post-Synaptic Potentials (PSPs) | No | No | **98.84** |
| *Lee et. al.* [15] | 3 layer H1=300, H2=300 | MSE and weight regularization | Numerical gradient computation performed on continuous valued Post-Synaptic Potentials (PSPs) | No | No | 98.77 |
| S4NN [8] | 2 layer H1=400 | MSE | Time to First Spike used for numerical gradient computation | No | No | 97.4 |
| Baseline ANN | 2 layer H1=1280 | softmax-CE and dropout | High precision conventional Backprop. | No | No | 98.67 |
| **This work** | **2 layer H1=1280** | **softmax-CE and dropout** | **Discrete spikes are used in IF neurons to implement all elements of Backprop. for learning** | **Yes** | **Yes** | **98.65 +/-0.04** |

Table 3: Generalization of SSNN-BP to EMNIST and FMNIST datasets

| Model | Network | Loss and Regularizer | EMNIST test-set accuracy (%) | FMNIST test-set accuracy (%) |
|---|---|---|---|---|
| HM2-BP [14] | 2 layer H1=800 | MSE | 85.41 | 88.99 |
| ST-RSBP [18] | 3 layer H1=400, H2=400, Recurrent Spiking Network | MSE | – | 90.13 |
| Baseline ANN | 2 layer H1=1280 | softmax-CE and dropout | **86 +/- 0.06** | **90.8 +/- 0.1** |
| This work | 2 layer H1=1280 | MSE | 81.26 +/- 0.04 | – |
| This work | 2 layer H1=1280 | MSE and dropout | 84.48 +/- 0.04 | – |
| **This work** | **2 layer H1=1280** | **softmax-CE and dropout** | **85.76 +/- 0.15** | **90.5 +/- 0.2** |

*Extended MNIST* (EMNIST) extends MNIST by adding lower and uppercase alphabets. We test our algorithm on the Balanced EMNIST dataset[34] which comprises of 131, 000 labeled grayscale image samples belonging to one of 47 classes. We use 120, 000 samples for training and test on 10, 000 samples from the remaining. The input encoding is identical to the one described for MNIST.

*Fashion MNIST*[33] is another dataset where 28x28 grayscale images of clothing items have to be classified into 10 classes, comprising of 60, 000 training and 10, 000 test samples. The input encoding described for MNIST is used.

The detailed performance comparisons are provided in Table 3. Our algorithm performs better on the EMNIST and Fashion MNIST dataset compared to state of the art in SNN learning[14], [18] and approaches ANN level of performance on both datasets.

Table 3 clearly demonstrates the improvement in performance achieved by dropout regularization as well as the approximate Softmax Cross-Entropy Loss function. Adding dropout improves the test classification accuracy on the EMNIST dataset from 81.3% to 84.5% and it is further improved to 85.8% by introducing the approximate Softmax Cross-Entropy Loss.

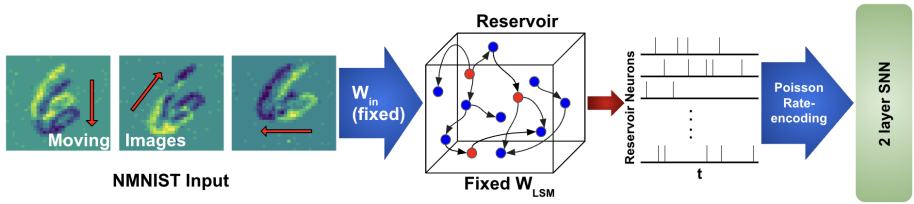

Figure 9: Setup for Neuromorphic-MNIST classification. A fixed spiking reservoir is used to perform a basic level of temporal processing on the input and is then classified using a 2-layer SNN trained with our method

3.3.3. *Naturally spiking input dataset - Neuromorphic-MNIST*

Neuromorphic-MNIST [35] is a neuromorphic dataset where MNIST images are recorded by an AER sensor programmed with fixed motions to generate spiking representations of the images. For our simulations, we use the standard

**ToFrame** preprocessing of the **Tonic** Python library [36] to convert the raw AER data of the N-MNIST samples into usable spike-trains. For any given N-MNIST AER spike, the preprocessing code first performs basic outlier detection by considering the number of spikes issued by neighboring neurons within a fixed time window and if it crosses a threshold, then the microsecond-level AER spike is extended into a millisecond-level frame pixel. This software-level preprocessing is functionally equivalent to pulse extender circuits, such as the ones described in [37] that are normally used to interface with AER sensors. Through this process, the approximately $300ms$ ($300,000\mu s$) long N-MNIST AER samples are converted into clocked N-MNIST samples, each containing approximately 300 frames of spikes for further processing. This is also unavoidable for practical simulations as it is both infeasible to perform SNN simulations with 300,000 timesteps (as would be necessary to maintain the AER timing precision) as well as unnecessary as the raw AER data is very sparse.

Since there is a temporal aspect to the data in N-MNIST, we use a fixed and randomly connected spiking Reservoir [38], [39] to perform a basic level of temporal processing before the resultant **spike rates** of the reservoir neurons are encoded by Poisson random spike-trains and input to a 2-layer SNN that is trained for the classification task by our algorithm.

In our algorithm, the spiking rates of the Reservoir neurons are recoded as Poisson spike trains. Recoding of the original reservoir neuron spike-trains into rate-based Poisson spike-trains needs to be done during the training phase to ensure that we generate stochastic spike-train inputs for the learning algorithm. For inference, we can directly use the reservoir spike trains as input without having to recode them as Poisson spike trains

In our simulations, we have computed the spike rate calculation after the Reservoir layer in two different ways:

**Ideal computation**: By counting the total number of spikes issued by each Reservoir neuron and normalizing it by the number of Simulation steps used for the LSM simulation ($\sum_t s_x^{Reservoir}(t)/T_s$)

**Realistic computation**: By convolving the Reservoir spike-trains with an exponential kernel and using a leaky integrator to estimate the instantaneous spike-rate - which is used to generate the random Poisson spike trains

The second method of recoding Reservoir neuron spike rates into Poisson spike trains outlined above would not require much overhead. It can be accomplished simply by using an additional layer of synapses (to implement the convolution with an exponential kernel) connecting each Reservoir neuron to a stochastic spiking neuron, [40] as shown in Fig. 10

Table 4 contains the comparisons of the performance of our method with the state-of-the-art in SNN learning. The equations and parameters for the LSM simulation are replicated from the basic 1000-neuron **LSM Model** described in [41]

Table 4: Generalization of SSNN-BP to N-MNIST dataset

| Model | Network | Loss and Regularizer | N-MNIST test-set accuracy (%) |
|---|---|---|---|
| **HM2-BP** [14] | 2 layer H1=800 | MSE | 98.84 |
| DECOLLE [42] | Convolutional Neural Network with 3 (7x7) Conv-Pool layers followed by a fully connected layer | MSE and dropout | 99 * (Results on a subset of the data with 20,000 training samples and 1000 test samples) |
| This work | 1000-neuron Reservoir + 2 layer SNN, H1=1280 | softmax-CE and dropout for 2-layer SNN only | 98.26 +/- 0.02 |

*3.4. Compatibility with Neuromorphic Hardware*

The supervised training of neural networks using backpropagation of errors is a data-intensive task. There has been a significant thrust to transfer these operations in/near-memory and on-chip for latency and power efficient forward and backward propagation [43]. Large neuromorphic cores of thousands of fast communicating neurons and energy efficient address event representation between hundreds of such cores have been demonstrated in digital and analog mixed mode hybrid design in hardware [44]. Chips like Neurogrid[43], TrueNorth[2], SpiNNaker[45], and Loihi[3] have made the realization of very large-scale neural networks possible, bringing down the time and energy complexity of network

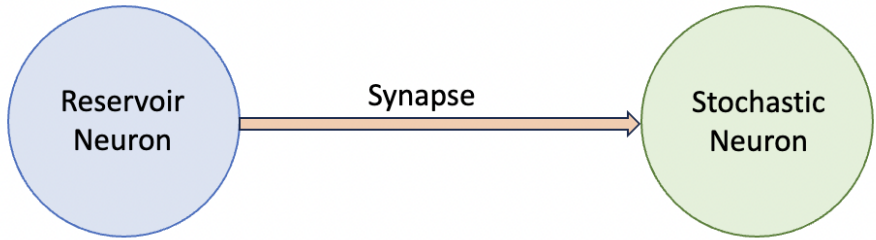

(a)

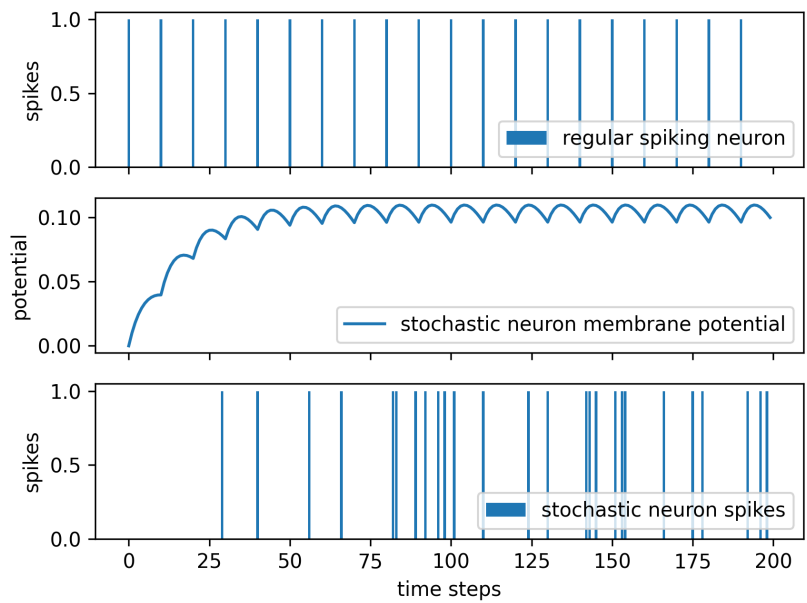

(b)

Figure 10: **(a)** Setup for recoding Reservoir neuron spike rates into Poisson spike trains using stochastic neurons **(b)** An example of the working of this setup - showing the spike rate of a regular spiking neuron being encoded as a Poisson random spike train

dynamics by several orders (3-4) of magnitude compared to conventional CPU and GPU simulations [44], [46].

The present trend in the neuromorphic VLSI is to increase in scale while maintaining the simplicity of individual units. Most implementations make use of simple leaky, integrate and fire neurons with methods to stochastically generate and propagate information only using spikes. These implementations also offer simple learning rules based on spike timings (STDP). The hardware acceleration benefits are not limited to just forward and backward propagation. Weight gradient calculation and weight update is another field where the role of hardware-enabled efficiency is extremely critical. As described earlier, the proposal of RPU enables simultaneous calculation of gradients and write-back in large crossbar arrays of analog memories holding network weights. Since RPUs depend on the generation of stochastic and pulsed representation of activations and errors, they are naturally well-suited for SNNs: spikes easily replace the stochastic pulses needed for weight update calculation to achieve the same effect when combined with spike coincidence-based learning rules.

However, due to the nature of RPU-based stochastic multiplication, this type of training algorithm is mainly relevant for clocked spiking systems where voltage pulses can align in time for predictable periods of time (i.e. a clock cycle). This is the case because we are using an RPU-based approach which requires pulse co-incidence to update the weights and is a common assumption for RPU-capable memories proposed as synapses, including analog memories [31], [23], [47], [48], [49]. As such, pulse-extender circuits, such as ones described in works such as [37], [50], [51] would be required to interface data from asynchronous AER sensors with our RPU-based SNN learning system

Several device-level implementations of stochastic neurons using resistive memories like phase change memories [52], bulk switching PCMO RRAMs [53] and nanomagnets [54] have been proposed. Similarly, analog memories for synapses like charge trap flash [23] and DRAM-like capacitors [47] with analog charge storage have shown the feasibility of RPU. Thus algorithms tailored to a generic description of an SNN and all computations done in spikes are readily transferable and scalable on neuromorphic hardware.

## 4. Conclusion

In this paper, we introduce a fully spike-based learning architecture for SNNs (SSNN-BP) that performs all computations using spiking neurons and performs weight updates using stochastic encoding to implement a spike coincidence mechanism, proposed for an in-memory on-chip learning implementation. The algorithm is spatially and temporally local, employs concurrent passes and incremental updates and does not require any external memory or numerical computation. We further show that its performance tends towards that of the rate-based version of the learning algorithm, for large $T_s = 300$, which performs on par with the state of the art in SNN learning algorithms on a number of benchmark classification tasks and approaches ANN level of performance. Furthermore, we show the performance improvement arising from introducing dropout regularization as well as adding lateral inhibitory connections in the output layer to approximate the Softmax Cross-Entropy loss function.